\documentclass[fleqn]{article}
\usepackage{epsfig}
\usepackage{epic}
\usepackage{ecltree}
\usepackage{isolatin1}
\usepackage{times}
\usepackage{titlesec}
\titleformat{\title}[hang]
{\fontsize{14}{16.8pt}\bfseries}{\thetile}{1em}{}

\titleformat{\abstract}[hang]
{\fontsize{9}{10.8pt}\bfseries}{\theabstract}{1em}{}

\titleformat{\section}[hang]
{\fontsize{12}{14.4pt}\bfseries}{\thesection}{1em}{}

\author{David Fernández-Amorós, Julio Gonzalo, Felisa Verdejo \\ Depto. de
  Lenguajes y Sistemas Informáticos, UNED \\ \{david,julio,felisa\}@lsi.uned.es} 
\date{}
\title{The Role of Conceptual Relations in Word Sense Disambiguation}
\begin{document}
\maketitle
\thispagestyle{empty}
\begin{abstract}
\leftmargin 1cm
\rightmargin 1cm
% Cambiar a 9pt Times New Roman con un centímetro de margen por cada lado
% en comparación con el texto normal
We explore many ways of using conceptual distance measures
in Word Sense Disambiguation, starting with the Agirre-Rigau
conceptual density measure. We use a generalized form of this
measure, introducing many (parameterized) refinements
and performing an exhaustive evaluation of all meaningful
combinations. We finally obtain a 42\% improvement over the original
algorithm, and show that measures of conceptual distance are not worse 
indicators for sense disambiguation than measures based on
word-coocurrence (exemplified by the Lesk algorithm). Our results,
however, reinforce the idea that only a combination of different
sources of knowledge might eventually lead to accurate word sense
disambiguation. 
\end{abstract}
\section{Introduction}
Competitive Word Sense Disambiguation (WSD) performance, as
illustrated by participants in the 
first Senseval competition~\cite{Kilgarriff-00}, can only be reached
mixing all kinds of knowledge: co-occurrence information, syntactic
information and collocations, additional information from dictionaries
such as domain labels, selectional restrictions, and all kinds of
heuristics (see for instance \cite{Ng-96,Rigau-97,Wilks-98b,
  Stevenson-99}). A problem with such hybrid systems is that they
make hard to discern what is the discriminative power of each of the
different types of knowledge about the context of  the word to be
disambiguate. Our belief is that a separate, detailed study of each
knowledge source is a necessary step to understand WSD challenges.

In this paper, we focus on conceptual relations as a source of
information for WSD systems. The basic hypothesis is that the right
senses for the words in a natural language expression will have closer 
semantically relations (in a semantic network) than incorrect 
combinations of senses. For instance, in ``{\em Spring is my
favorite
season}'', the {\em springtime} sense of spring has a hyponymy 
(IS-A) relation with the {\em season of the year} meaning of season,
while any other combination of senses (e.g. \emph{spring} as fountain and
\emph{season} as sports season) have weaker semantic relationships. 

Our aim is to perform an in-depth study (via exhaustive empirical
evaluation) of the role that conceptual relations may play in accurate
WSD. As a point of departure, we chose one of the most promising WSD
methods based solely on conceptual relations, the Agirre-Rigau
algorithm, based on a measure of conceptual
density~\cite{Agirre-96}. As in their work, we have used the
WordNet~\cite{Fellbaum-98} semantic network as the lexical database
providing word senses and semantic relations between them. Wordnet
includes around 168000 English word senses, and has also large-scale
versions for many other languages~\cite{Vossen-98}.

Then we generalized the original algorithm, parameterizing many
aspects of the original system, including the conceptual density
formula itself. The strategies incorporated to the
algorithm include as much possibilities to exploit 
semantic relations as we could think of. Finally, we performed an
exhaustive evaluation, running the system in more than 50 different
configurations against all nouns in the Semcor test collection, the
largest semantically annotated test collection known to us (even the
original algorithm had not been previously tested against the whole Semcor
collection).

In Section~2, the main algorithm and all the variants are
explained. Section~3 describes the evaluation performed and the
results obtained. Finally, Section~4 describes the main conclusions.
%%%%%%%%%%%%%%%%%%%%%%%%%%%%%%%%%%%%%%%%%%%%%%%%%%%%%%%%%%%%%%%
\section{Description of the algorithm}
The basic elements for the algorithm are a Lexical Knowledge Base
(LKB) with conceptual information (such as Wordnet {\em synsets}, or
sets of synonym terms), a binary relation 
\emph{R} (usually the hypernymy relation, equivalent to the IS-A
relation in an ontology) between the concepts in the LKB 
and a conceptual density formula (see below) giving the conceptual
density of a concept with a certain amount of \emph{activated} (with
respect to \emph{R}) subconcepts.

To disambiguate a word we do the following: first, we take the
surrounding text and form a window with a given fixed radius. Then we
rank the senses of the central word following these steps: 

\begin{itemize}
\item We look up the
senses of all words in the window. For every sense of every word, we
take a number of related concepts via the \emph{R} relation, and we 
weight them according to some formula. 
\item For each sense of the central word, the concept (related to the
  sense via transitive application of $R$) that has a highest
  conceptual density gives the conceptual density of the sense. 
\item Then we normalize the ranks for the senses of the word, and take
  the resulting values as output of the algorithm.
\end{itemize}

These steps define a conceptual density algorithm ``template'' with a
wide range of possibilities. In the next section we discuss the
parameters we have considered and the values we have tested.

\subsection{Parameters}

\begin{description} % Parametros

\item[Transitive relation $R$.] The most obvious is perhaps the
  hypernymy relation, but we have also considered the union of 
  semantic relations such as hypernymy and meronymy (``PART-OF''
  relation). 

\item[Conceptual density measure.] We have tested four different conceptual density measures:
\begin{enumerate} % Formulas
\item The original Agirre-Rigau conceptual density
  formula~\cite{Agirre-96}: 
\begin{equation}
CD(c,m) = \frac{\displaystyle \sum_{i=0}^{m-1}adesc^{i^{\alpha}}}
{\displaystyle \sum_{i=0}^{h-1}adesc^i}
\end{equation}
where \emph{adesc} is the average number of descendants of concept $c$
according to $R^{*^{-1}}$, $\alpha=0.2$, $m$ is the number of marks in
the subhierarchy of $c$. And $h$ is the depth of the subhierarchy under
$c$. We have called this formula Strict Agirre-Rigau (SAR). The $\alpha=0.2$
value optimizes the results for WordNet1.4. 

\item The same formula without $\alpha$ (which was optimized by Agirre 
  and Rigau for a different, much smaller test collection).

\item The \emph{simple density formula}(SDF) = $m$ / $desc_c$. A
  simple baseline to test the importance of the conceptual density
  formula. 
\item The \emph{logarithmic formula} (LF) $=  \frac{1}{desc_c}
  log_2{d} \displaystyle \sum_{i=0}^{m-1}adesc^{i}$. Where $d$ is the
  depth of the concept $c$ in the hierarchy. This is the AR formula
  with a correction factor to favor more specific concepts (deeper in
  the hierarchy) 
\end{enumerate} % Formulas

\item[Window size.] We have experimented with various window sizes.

\item[Selection of related concepts] When it comes to selecting the
  concepts related to a sense through \emph{R} we have taken several
  possibilities into account.  

\begin{itemize} % Top y level

\item First, we have a parameter to rule out the upmost levels of
  the hierarchy induced by the
  transitive closure of \emph{R}. The reason for this is that the
  higher levels in broad conceptual hierarchies 
  tend to be helplessly subjective. If there is a concept
  representative of the topic discussed in the word window and this
  concept is supposed to be meaningful for the disambiguation task, it
  shouldn't be too abstract or generic (as the concepts at higher levels
  usually are) in relation to the word senses being disambiguated. The
  conceptual density formulas are designed to reflect this but with
  big window sizes it is inevitable that WordNet tops get high
  densities. A value of 0 in this parameter represents considering
  the whole hierarchy.  

\item
We introduce another parameter \emph{l}, to consider
only the nearer \emph{l} concepts through transitive
application of \emph{R}. In other words, when computing the conceptual
density of a concept \emph{c}, we won't consider the weight of a
subconcept \emph{s} if we have to iterate through \emph{R} more than
\emph{l} times to reach \emph{c}. The idea behind this is that a
concept \emph{c} and its immediate hypernym will be closely related
semantically (as would be the case between
highway\_1\footnote{Following the convention that w\_i is the ith
  sense in WordNet of the word \emph{w}}and road\_1 in
Wordnet); On the contrary, although a highway is surely an entity it
is unclear that this 
information has any semantic impact in a disambiguation task. It may
seem that this parameter and the one described above 
will yield similar results, but they show a very different behavior in 
our experiments. In 
this parameter, a limit value of 0 represents taking into
account all the concepts related via \emph{R}.

\end{itemize} % Top y level

\item[Sense weighting.] To compute the conceptual density of a concept \emph{c}, in the
  hierarchy induced by \emph{R} we have considered three possibilities
  to count and weight how many marks, \emph{m}, lie below : 

\begin{description} % Pesado

\item[synsets]
Counting each sense of the words in the window related to \emph{c} as
one mark. This is the original formulation of
Agirre \& Rigau. The problem here is that the words
interfere severely with themselves. If we take for instance the word
{\em end}, which has 14 senses in WordNet, and draw the hypernymy
hierarchy for the senses under {\em entity} (with some intermediate
nodes omitted for clarity) we get the results in Figure~\ref{tree}. 

\begin{figure}[htbp]
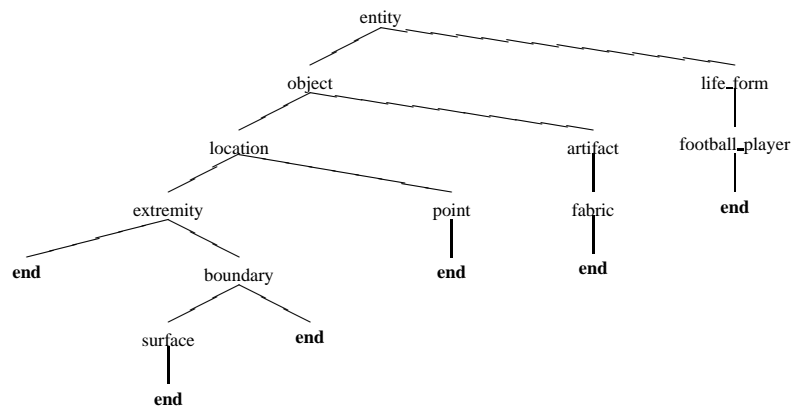

{\scriptsize
  \begin{center}
\setlength{\GapDepth}{5mm}
\setlength{\GapWidth}{15mm}
%\begin{picture}%(10,10)(-30,55)
  \begin{bundle}{entity}                           % entity
    \chunk{
      \begin{bundle}{object}                  % entity -> object
        \chunk{
          \begin{bundle}{location}            % entity -> object -> location
            \chunk{
              \begin{bundle}{extremity}   % entity -> object -> location -> region -> ex
                \chunk{\textbf{end}}               % entity -> object -> location -> region -> ex
                    \chunk{
                      \begin{bundle}{boundary}  % ent -> obj -> loc -> reg -> ex -> boun
                        \chunk{
                          \begin{bundle}{surface}  % ent -> obj -> loc -> reg -> ex->boun...
                            \chunk{\textbf{end}}
                          \end{bundle}
                          } %  surface // ent -> obj -> loc -> reg -> ex -> boun 
                      \chunk{\textbf{end}}
                      \end{bundle}
                      }    %  boundary // ent -> obj -> loc -> reg -> ex 
                  \end{bundle}
                  }     %   extremity // ent -> obj -> loc -> reg 
                \chunk{\begin{bundle}{point}  %     ent -> obj -> loc -> point             
                    \chunk{\textbf{end}}
                  \end{bundle}
                  }       % point //  ent -> obj -> loc 
              \end{bundle}
              }         % location // ent -> obj
            \chunk{\begin{bundle}{artifact}   %  //    ent -> obj -> arti     
       
                \chunk{\begin{bundle}{fabric} %       ent -> obj -> arti -> fabric        
                    \chunk{\textbf{end}}
                  \end{bundle}
                  } % fabric   //  ent -> obj -> arti 
              \end{bundle}
              }   % artifact //  ent -> obj
          \end{bundle}
          }     % object   //  ent
        \chunk{
          \begin{bundle}{life\_form}   % ent -> life_f               
            \chunk{
              \begin{bundle}{football\_player} % ent -> life_f -> footb_pl    
 
                \chunk{\textbf{end}}
              \end{bundle}
            }   % football_player
        \end{bundle}
        }     % life_form
    \end{bundle}       % entity

%\end{picture}
    \caption{the hierarchy of end}
    \label{tree}
  \end{center}
}
\end{figure}

It is easy to see here that the remaining 8 senses of end (which are not
hyponyms of entity) will probably be 
discriminated against these because, in the absence of any context, the
concept \emph{object} in the figure gets a very high density. If we
add more words as 
context in the window, the chances are that the majority of the senses
will fall under the subhierarchy of \emph{entity} and the algorithm would
discard the other senses. Another adverse effect of highly polysemous
words is that they tend to dominate the conceptual density
measures. For instance, \emph{end} has 14 senses and therefore 
14 marks in the density measures, and that seems very unfair given that
around one-third of the words in running text are
monosemous. In order
to minimize these effects, we have tested two additional forms of weighting
senses:

\item[fractional]

Counting for every sense a word in the window $1/m$ (where m is the
 total number of senses of that word) to prevent a highly polysemous
 word from biasing the conceptual density, although probably this won't
 prevent some words from disambiguating themselves.  
\item[words]
Counting as marks under the subhierarchy of a concept only the number
 of different words in the window contributing with senses under
 \emph{c}. This way, all words in the window will contribute the same and
 also a high local in-word density (usually derived of the
 fine-grainedness of WordNet) shouldn't discriminate the senses of
 that word outside that area. 
\end{description} % Pesado

\end{description}   % Parametros
\section{Evaluation}
The evaluation has been conducted on the Semcor
collection~\cite{Miller-93}, a set of 171 documents where all content
words are annotated with the most appropriate Wordnet sense. In our
evaluation, each of the versions of the WSD algorithm has been tested
on every noun in every Semcor document. 

The behavior of the system is reported as a {\em recall} measure as
defined for the first SENSEVAL campaign~:

The score regime allows scores between 0 and 1 where the system
returns more than one sense for an instance, with the probability mass
shared. Recall is computed by dividing  the system's scores over all
correct senses by the total number of items to be disambiguated.

This measure compares correct disambiguations against all nouns in the 
collection; therefore, a system that is very precise but has a low
coverage will also have a low recall overall. 

\subsection{Overall performance}

Table~\ref{overall} compares the original Agirre-Rigau algorithm, our
best conceptual density system, and three reference measures: a most
frequent sense heuristic (always picking up the first wordnet sense), a
random WSD baseline and a classical WSD strategy based on coocurrence
of words in dictionary definitions (Lesk).

\begin{table}[htbp]
  \begin{center}
    \begin{tabular}{ll}
      WSD algorithm & Recall \\[.5em]
      \hline\\
      UNED conceptual density & 31.3\% \\
      Lesk  & 27.4\% \\
      Agirre-Rigau & 22.0\% \\[.5em]
      \hline\\
      Random baseline & 28.5\% \\
      Most frequent heuristic & 70.0\% \\
    \end{tabular}
    \caption{Overall performance}
    \label{overall}
  \end{center}
\end{table}

Surprisingly, the recall of the original Agirre-Rigau system is below
the random baseline. This figure is slightly misleading, because the
precision of Agirre-Rigau is above the random baseline; But a
random election has a 100\% coverage, while the original conceptual
density measure is not able to disambiguate all words. In any case,
the performance of the original density measure is much poorer than
expected. Results reported in \cite{Agirre-96} were more promising,
but they were obtained on a test collection 50 times smaller than the
whole Semcor collection (they used only four Semcor documents).

Our best system achieves 31.3\% recall, a 42\% improvement over the
original Agirre-Rigau system. This is a dramatic improvement with
respect to the
original algorithm, but still the results are far below the most
frequent sense heuristic. The comparison with the 70\% recall of this
simple heuristic could lead to discard conceptual relations as a
source of information for the Word-Sense Disambiguation task. This
would be, however, an erroneous conclusion, for a number of reasons:

\begin{itemize}
\item We have also compared the performance of conceptual density with 
  a classical WSD algorithm based on contextual information in
  dictionary definitions~\cite{Lesk-86}, which was used as a strong
  baseline in SENSEVAL 1~\cite{Kilgarriff-00}. The recall of Lesk algorithm
  is 27.4\%, also below the random baseline! There are some reasons
  that explain these results: 
  \begin{itemize}
  \item The human annotations, taken as a gold
    standard, are biased in favor of the first wordnet sense,
    (which corresponds to the most frequent). Human annotators, in an
    all-words disambiguation task, have to select the appropriate
    sense for a different word in each iteration, each word having
    more than 5 senses in average. Inevitably, the annotator tends to
    pick up the first sense that seems to fit the context, and this
    produces a bias in favor of higher ranked senses. Studies on WSD
    evaluation \cite{Resnik-99,Kilgarriff-00} have argued in favor of
    a lexical sample task, where the annotator repeatedly annotates
    occurrences of the same word, reaching a minimal familiarity with
    the senses of the chosen word. This was the approach taken in
    SENSEVAL-1, where the Lesk algorithm behaves much better than in
    this Semcor-based evaluation. Unfortunately, the SENSEVAL test
    collection is based on a different dictionary ({\em Hector}), and
    thus cannot be used to test our conceptual density strategies.
  \item Beyond human annotation problems, the all-words task implies
    that the system must be repeatedly attempting to disambiguate
    instances of very common terms, 
    which may have 20 different senses in the database. This terms are 
    almost impossible to disambiguate, and probably also useless to
    disambiguate for a majority of applications.
  \end{itemize}
A more appropriate conclusion would be, then, that neither conceptual
nor contextual measures are sufficient, in  isolation, to perform
accurate Word Sense Disambiguation. 
\item Our algorithm assigns probabilities to senses (unlike the most
  frequent sense heuristic), and the overall distribution of
  probabilities produces better results in a Text Retrieval system
  based on concept retrieval than the most frequent heuristic, as we
  have previously reported in~\cite{siglex-99}. This is an indication
  that the recall measure in a pure WSD task may not reflect the
  utility of a WSD system in final Natural Language applications. This 
  indirect measure is a proof of the potential utility of conceptual
  density measures for Word-Sense Disambiguation.
\end{itemize}

We will focus now on the separate evaluation of all the variants
introduced in the original algorithm, which led us to the best
combination reported here.

\subsection{Type of conceptual relation}

Table~\ref{relations} shows the results of the algorithm using
different types of semantic relations. Apparently, meronymy/holonymy
relations do not add any useful information to hypernymy.
\begin{table}[htbp]
  \begin{center}
    \begin{tabular}{||l|c||} \hline
      Relation & Recall\\ \hline
      Hypernymy                & 31.28\% \\ \hline
      Hypernymy + Meronymy     & 31.28\% \\ \hline
      Hypernymy + Holonymy     & 30.88\% \\ \hline
      Meronymy                 & 26.62\% \\ \hline
      Holonymy                 & 27.00\% \\ \hline
    \end{tabular}
    \caption{Recall with different conceptual relations}
    \label{relations}
  \end{center}
\end{table}

\begin{figure}[ht]
  \begin{center}
    \epsfig{file=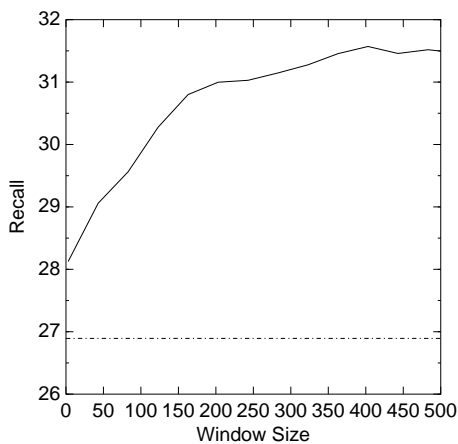, width=6cm}
    \caption{Effects of window size}
    \label{window-size}
  \end{center}
\end{figure}

\subsection{Window size}

Figure~\ref{window-size} shows the behavior of the algorithm with
window sizes between 1 and 500 words. Remarkably, disambiguation gets
consistently (although steadily) better with larger window sizes up to
150 words. This probably means the contextual information in a whole
document is useful to disambiguate a word, providing topic
information for the document.

\subsection{Conceptual Density Formula}

The effects of the density formula can be seen in
Table~\ref{formula}. The alternative formulations LF and SDF behave
worse than the original formula by Agirre and Rigau. However, their
$\alpha$ parameter, which was adjusted to $0.2$ in order to optimize
disambiguation over four particular Semcor documents in WordNet 1.4, is clearly
inadequate when evaluating against all Semcor documents in WordNet 1.5~: 
$\alpha=1$ (AR) produces a $10$\% improvement against $\alpha=0.4$
(SAR).  
\begin{table*}[htbp]
  \begin{center}
    \caption{Effects of conceptual density measures}
    \begin{tabular}{||l|c||} \hline
      Density Formula & Recall\\ \hline
      AR              & 31.28\% \\ \hline
      SAR             & 27.45\% \\ \hline
      LF              & 26.60\% \\ \hline
      SDF             & 23.29\% \\ \hline
    \end{tabular}
    \label{formula}
  \end{center}
\end{table*}

The different formulas give recall figures between 23.3\% and 31.3\%
(50\% greater), showing that choosing an adequate formula has a direct 
impact on the results. Perhaps a better formula could further improve 
results.

\begin{figure}[hbp]
  \begin{center}
    \epsfig{file=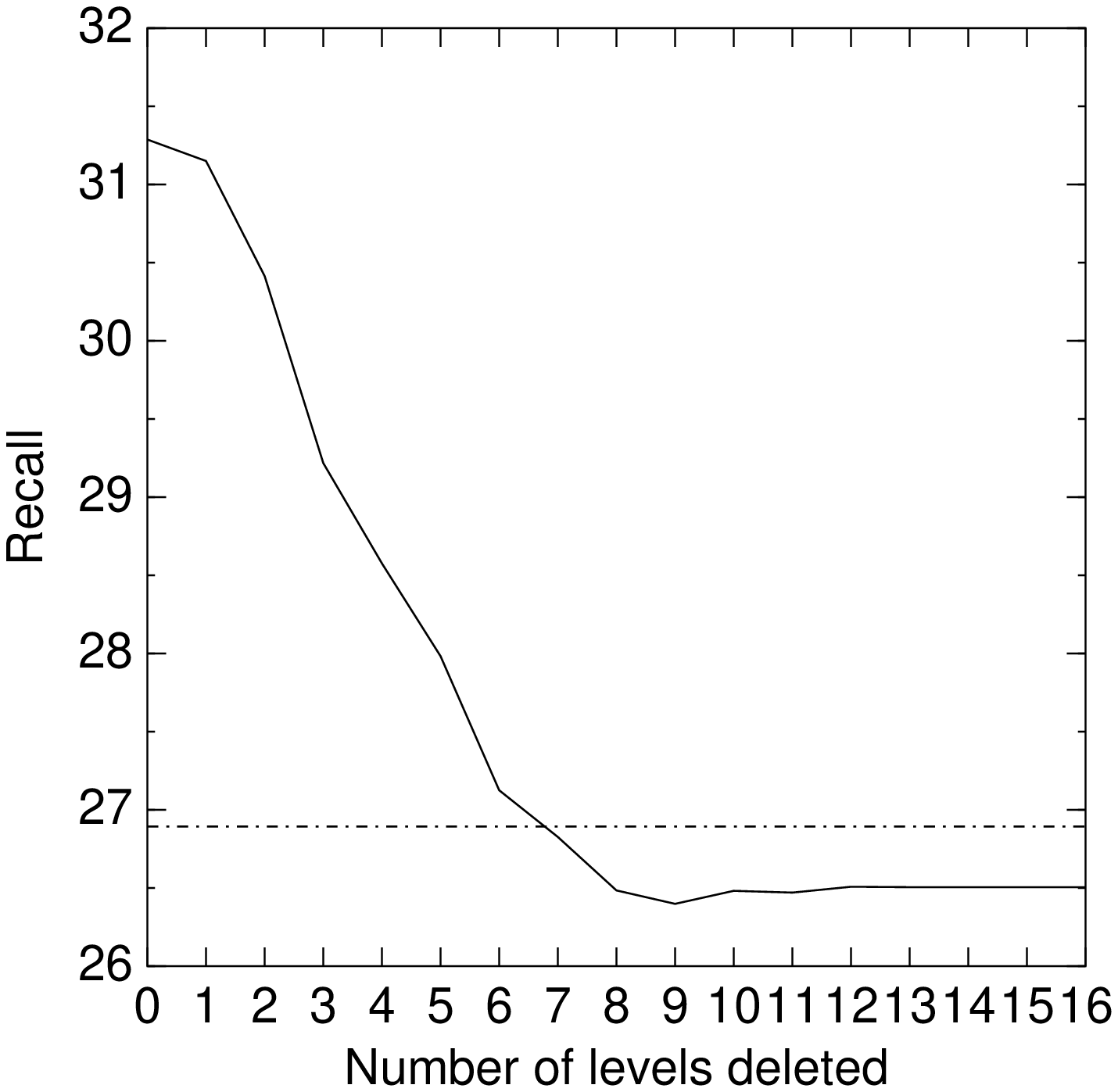, width=6cm}
    \epsfig{file=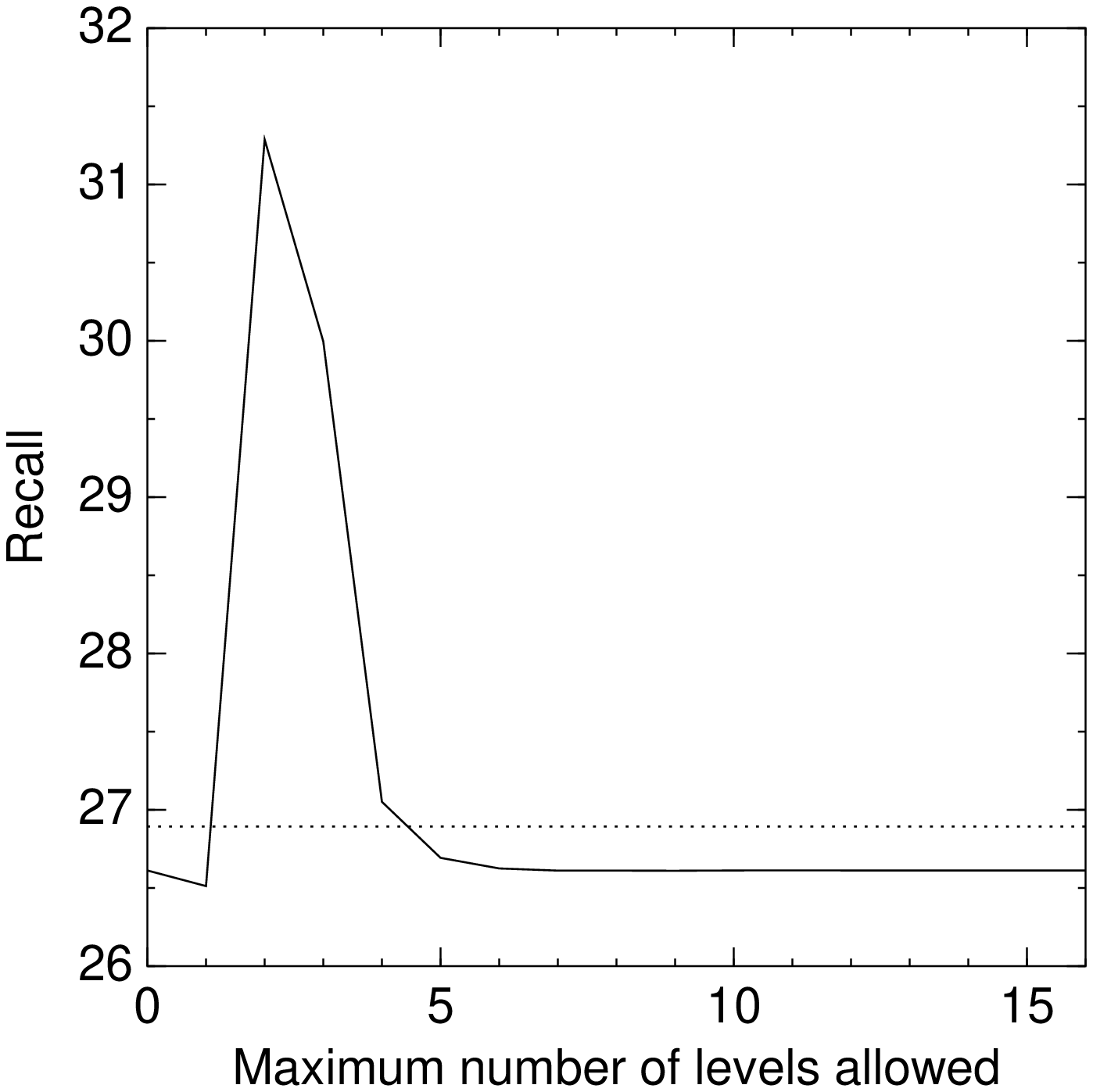, width=6cm}
     \caption{Selection of synsets}
   \label{upperlevels}
  \end{center}
\end{figure}

\subsection{Selection of synsets}

\subsubsection*{Removal of upper levels}

Figure~\ref{upperlevels} (left plot) shows the effects of removing upper levels of 
the hierarchy. Contrary to our hypothesis, even only removing the two
upper levels harms the recall of the system. Removing more than 6
levels produces a random behavior, as most information in WordNet lies
in the first 6 levels. 
This results seems to indicate that the WSD algorithm is not
performing as expected: the upper levels are used in the
disambiguation, and therefore the conceptual density measure is using
conceptual relationships that are far too general to be meaningful for 
disambiguation. This can partially explained the poor absolute
performance of the density measures.

\subsubsection*{Upper limit on hierarchical chains}

The effects of limiting the inspection of hypernym chains are shown
in Figure~\ref{upperlevels} (right plot). The plot shows that the algorithm is
useless without such limitation, and the optimal limit is two.

This criterion confirms that going up in the hierarchy without
limitation introduces noise - due to underspecific concepts - that
spoils the performance of the algorithm.

\subsection{Sense weighting}

Table~\ref{weighting} shows recall for the three approaches to sense
weighting. Surprisingly, assigning lower weights to senses for highly
ambiguous words ("Fractional") does not improve performance over the
standard approach ("Synsets"). Taking the number of different source
words in the density measure ("Words") produces an improvement, but
nearly negligible. 

\begin{table}[htbp]
  \begin{center}
    \begin{tabular}{||l|c||} \hline
      Criterion & Recall\\ \hline
      Words       & 31.28\% \\ \hline
      Synsets     & 30.81\% \\ \hline
      Fractional  & 27.94\% \\ \hline
    \end{tabular}
    \caption{Effects of sense weighting}
    \label{weighting}
  \end{center}
\end{table}

\subsection{Behavior on different text categories}

The Semcor documents, a fraction of the Brown Corpus~\cite{Francis-82}, 
are classified according to a set of predefined domains (Press,
General Fiction, Romance, Humor, etc.). It is interesting to see how
WSD performance varies along different document categories. In
Table~\ref{categories}, overall performance is split according to
such categories. Categories where conceptual density works better are
ranked higher in the table.

\begin{table}[htbp]
  \begin{center}
    \begin{tabular}{||l|r|r|r||} \hline
      Text category  & Random recall & Algorithm recall & Improvement\\ \hline
      A. Press: reportage                  & 26.95\% & 36.67\% & 36.08\% \\ \hline
      C. Press: reviews                    & 27.06\% & 34.91\% & 29.01\% \\ \hline
      E. Skills \& hobbies                 & 26.89\% & 33.68\% & 25.26\% \\ \hline
      F. Popular Lore                      & 26.73\% & 32.79\% & 22.66\% \\ \hline
      D. Religion                          & 25.75\% & 31.22\% & 21.23\% \\ \hline
      H. Miscellaneous                     & 26.14\% & 31.65\% & 21.09\% \\ \hline
      J. Learned                           & 27.17\% & 32.78\% & 20.64\% \\ \hline
      L. Mystery \& detective fiction      & 25.29\% & 29.89\% & 18.16\% \\ \hline
      P. Romance \& love story             & 25.03\% & 29.19\% & 16.63\% \\ \hline
      B. Press: editorial                  & 27.64\% & 31.69\% & 14.65\% \\ \hline
      G. Belles lettres, biography, essays & 28.14\% & 31.93\% & 13.48\% \\ \hline
      M. Science fiction                   & 26.49\% & 29.87\% & 12.74\% \\ \hline
      K. General fiction                   & 25.63\% & 28.32\% & 10.49\% \\ \hline
      R. Humor                             & 26.66\% & 29.06\% & 9.00\% \\ \hline
      N. Adventure \& western fiction      & 22.08\% & 23.06\% & 4.45\% \\ \hline
    \end{tabular}
    \caption{WSD performance in different text categories}
    \label{categories}
  \end{center}
\end{table}

The results are remarkable. While a random disambiguation produces
similar recall figures (indicating that the mean polysemy is similar
in any kind of documents), the WSD system performs better on non-fiction
categories ({\em Press: reportage, reviews, skills and hobbies},
etc.), and worse on fiction categories ({\em adventure, humor, general
fiction}). Conceptual density improves random WSD in a 36\% for {\em
Press: reportage}, while for {\em Adventure \& Western Fiction} the
improvement is negligible (4.5\%). This confirms the hypothesis that
WSD is more plausible in technical documents, where word senses have
clearer distinctions, metaphors are less common, and the context
provides more accurate domain information. 

\section{Conclusions}
We have provided an exhaustive evaluation of different WSD algorithms
that rely solely on the conceptual relations between candidate word
senses. Our point of departure has been the Agirre-Rigau algorithm,
based on a conceptual density measure over the WordNet hierarchy. This 
algorithm, which had a competitive performance over smaller test
collection, behaves poorly in a complete evaluation against all
Semcor documents. We have experimented with many kinds of
improvements to the algorithm, and tuned all parameters associated to
them, obtaining evaluations for more than fifty variants of the WSD
system. For comparison purposes, we have also implemented and
evaluated a simple version of the classical Lesk algorithm, based
solely on contextual information from dictionary definitions, which is 
also used in SENSEVAL WSD competitions.

Some of the main conclusions from our experiments are:

\begin{itemize}
\item Our best system performs 42\% better than the original
  Agirre-Rigau algorithm, and 14\% better than the Lesk
  algorithm based on coocurrence in dictionary definitions. This
  improvement is obtained with an implementation that runs in linear
  time and has been used to disambiguate large text collections in
  three different languages (English, Spanish and Catalan) within the
  ITEM project~\cite{lrec-00}. 
 Its performance, however, is still low in terms of absolute
  recall, indicating that conceptual relations should be combined
  with other types of information (contextual, syntactic, domain
  information, etc.). We have also argued that the test collection
  itself -Semcor- is not appropriate for testing systems; it is
  desirable that the new Senseval initiative creates a better
  evaluation ground to provide a reliable way of measuring the
  effectiveness of WSD systems in final NLP tasks.
\item We have shown that, in practice, the original Agirre-Rigau
  algorithm uses long hierarchical chains to disambiguate, which are
  associated with vague conceptual associations that give noisy
  results. Our optimal setting uses maximal hypernymy chains of size
  2, combined with other optimizations to keep the coverage of the
  system. We have also shown that bigger window sizes provide better
  results, because they exploit all domain information in a text to
  disambiguate. 
\item We have provided quantitative evidence proving that WSD is more
  feasible on non-fiction, domain-specific documents rather than on
  general fiction texts with this technique. 
\item Finally, we have shown that a direct comparison of recall with a 
  Most frequent heuristic over Semcor does not reflect the properties
  of WSD systems; our system has a much lower recall than this
  heuristic, but gives better results in a text retrieval experiment
  using word senses, as we have previously shown in \cite{siglex-99}.
  This is a strong evidence that the evaluation of 
  WSD systems should also be measured indirectly in NLP applications. 
\end{itemize}

\section{Acknowledgments}

This work has been supported by a Ph.D. research grant from the
Comunidad Autónoma de Madrid, and also by the Spanish Comisión
Interministerial de Ciencia y Tecnología, project {\em Hermes}
(TIC2000-0335-C03-01). 

%%% Local Variables: 
%%% mode: latex
%%% TeX-master: t
%%% End: 

\bibliographystyle{alpha}
\bibliography{ir}

\newcommand{\etalchar}[1]{$^{#1}$}
\begin{thebibliography}{VGP{\etalchar{+}}00}

\bibitem[AR96]{Agirre-96}
Eneko Agirre and German Rigau.
\newblock Word sense disambiguation using conceptual density.
\newblock {\em COLING (International Conference in computational linguistics)},
  1996.

\bibitem[Fel98]{Fellbaum-98}
C.~Fellbaum.
\newblock A semantic network of english: the mother of all wordnets.
\newblock {\em Computers and the Humanities, Special Issue on
  \uppercase{E}uroWordNet}, 1998.

\bibitem[FK82]{Francis-82}
W.~N. Francis and H.~Kucera.
\newblock {\em Frequency Analysis of English Usage: Lexicon and Grammar}.
\newblock Houghton Mifflin Company, Boston, 1982.

\bibitem[GRA97]{Rigau-97}
Jordi~Atserias German~Rigau and Eneko Agirre.
\newblock Combinig unsupervised lexical knowledge methods for word sense
  disambiguation.
\newblock In {\em Proceedings of the ACL (Association for Computational
  Linguistics)}, 1997.

\bibitem[KR00]{Kilgarriff-00}
A.~Kilgarriff and J.~Rosenzweig.
\newblock Framework and results for english \uppercase{SENSEVAL}.
\newblock {\em Computers and the Humanities}, 34(1-2), 2000.

\bibitem[Les86]{Lesk-86}
Michael~E. Lesk.
\newblock Automatic sense disambiguation using machine readable dictionaries~:
  How to tell a pine cone from an ice cream cone.
\newblock In {\em Proceedings of SIGDOC (Special Interest Group for
  Documentation) Conference, Toronto, Canada}, 1986.

\bibitem[MLTB93]{Miller-93}
G.~A. Miller, C.~Leacock, R.~Tengi, and R.~T. Bunker.
\newblock A semantic concordance.
\newblock In {\em Proceedings of the ARPA (Advanced Research Projects Agency)
  Workshop on Human Language Technology}. Morgan Kauffman, 1993.

\bibitem[NL96]{Ng-96}
H.~Ng and H.~Lee.
\newblock Integrating multiple knowledge sources to disambiguate word sense: an
  exemplar-based approach.
\newblock In {\em Proceedings of ACL (Association for Computational
  Linguistics)}, 1996.

\bibitem[RY99]{Resnik-99}
P.~Resnik and D.~Yarowsky.
\newblock Distinguishing systems and distinguishing senses: New evaluation
  methods for word sense disambiguation 5(2):113--134.
\newblock {\em Natural Language Engineering}, 1999.

\bibitem[SW99]{Stevenson-99}
Mark Stevenson and Yorick Wilks.
\newblock Combining weak knowledge sources for sense disambiguation.
\newblock {\em Proceedings of IJCAI (International Joint Conference on
  Artificial Intelligence) in Stockholm, Sweden}, 1999.

\bibitem[VGP{\etalchar{+}}00]{lrec-00}
F.~Verdejo, J.~Gonzalo, A.~Peñas, F.~López, and D.~Fernández.
\newblock Evaluating wordnets in \uppercase{C}ross-\uppercase{L}anguage
  \uppercase{T}ext \uppercase{R}etrieval: the \uppercase{ITEM} multilingual
  search engine.
\newblock In {\em Proceedings LREC (International Conference on Language
  Resources and Evaluation)}, 2000.

\bibitem[Vos98]{Vossen-98}
P.~Vossen.
\newblock {\em \uppercase{E}uro\uppercase{W}ord\uppercase{N}et: a multilingual
  database with lexical semantic networks}.
\newblock Kluwer Academic Publishers, 1998.

\bibitem[VPG99]{siglex-99}
P.~Vossen, W.~Peters, and J.~Gonzalo.
\newblock Towards a universal index of meaning.
\newblock In {\em Proceedings of SIGLEX(Special Interest Group on the
  Lexicon)}, 1999.

\bibitem[WS98]{Wilks-98b}
Yorick Wilks and Mark Stevenson.
\newblock Word sense disambiguation using optimised combinations of knowledge
  sources.
\newblock {\em Proceedings of COLING (International Conference in computational
  linguistics) - ACL (Association for Computational Linguistics) in Montreal,
  Quebec, Canada}, 1998.

\end{thebibliography}
\end{document}